\pdfoutput=1

\documentclass[11pt]{article}

\usepackage[final]{acl}

\usepackage{times}
\usepackage{latexsym}

\usepackage[T1]{fontenc}

\usepackage[utf8]{inputenc}

\usepackage{microtype}

\usepackage{inconsolata}

\usepackage{graphicx}
\usepackage{todonotes}

\title{Multimodal Contextual Dialogue Breakdown Detection for Conversational AI Models}

\author{Md Messal Monem Miah, Ulrike Schnaithmann, Arushi Raghuvanshi \textnormal{and} Youngseo Son \\
  Infinitus Systems, Inc. \\
  \texttt{\{messal.miah, ulie.schnaithmann, arushi, youngseo.son\}@infinitus.ai} }

\begin{document}
\maketitle
\begin{abstract}
Detecting dialogue breakdown in real time is critical for conversational AI systems, because it enables taking corrective action to successfully complete a task. In spoken dialogue systems, this breakdown can be caused by a variety of unexpected situations including high levels of background noise, causing STT mistranscriptions, or unexpected user flows.
In particular, industry settings like healthcare, require high precision and high flexibility to navigate differently based on the conversation history and dialogue states. This makes it both more challenging and more critical to accurately detect dialogue breakdown. To accurately detect breakdown, we found it requires processing audio inputs along with downstream NLP  model inferences on transcribed text in real time. In this paper, we introduce a Multimodal Contextual Dialogue Breakdown (MultConDB) model. This model significantly outperforms other known best models by achieving an F1 of 69.27.
\end{abstract}

\section{Introduction}

Dialogue breakdown detection is important in industry settings, because it allows the system to correct for mistakes in real time. While it is even better to avoid dialogue breakdown to begin with, in many industry settings there are components of the pipeline that have noise in real world settings. For example, the vendor or system making a voice call could drop some audio packets. The ASR vendor or model could miss some transcripts or have very noisy transcripts, especially when the user is in a setting with a lot of background noise or using a phone line with a poor network. With dialogue breakdown, we can detect that there was likely some missing context and say something like “Sorry I missed that, could you repeat yourself?” to get the conversation back on track.

\begin{figure}
    \centering
    \small
    \includegraphics[width=3in]{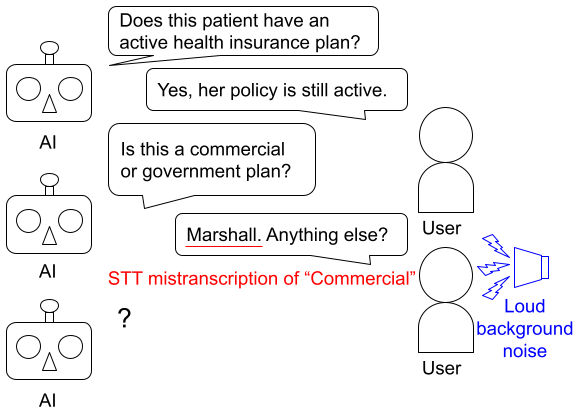}
    \caption{Example of dialogue breakdown in a phone call conversation caused by loud noise from user audio. See more examples in Section~\ref{sec:db_example} 
    }
    \label{fig:dialogue_breakdown_example}
\end{figure}

While dialogue breakdown is a challenging problem in general, there are some unique challenges in industry settings. In professional settings, users do not use as much explicit language or profanities. Instead of detecting this strong language, we often need to rely more on tone or cadence to detect user frustration. Additionally, there is low tolerance for incorrect responses. For example, in the healthcare domain a failed conversation could affect the time it takes for a patient to receive treatment. Finally, some industry use cases, including ours, have very complex and varied flows. For example, the average conversation in our domain consists of about 100 turns and context from early in the conversation can affect the flow even at the end of the conversation.

There are additionally unique challenges for detecting dialogue breakdown in phone call settings. Over the phone, there are strict latency requirements (e.g. delayed or repeatedly incorrect responses can cause frustration or even hang ups from users interacting with the system). In contrast, text-based chatbot systems often have visual feedback to indicate processing time and often target users and use cases with more leniency around latency and potential hallucinations. Thus, detecting dialogue breakdowns in a timely manner is crucial for real-time conversational speech AI systems. It is an extremely challenging task because there are multimodal factors in different components of the system pipeline  which can appear as diverse downstream issues. 

We found that prior state of the art models were not able to accurately capture dialogue breakdown in our industry setting. In this paper, we propose a new model which uses audio and text signals to predict dialogue breakdown generalizable to various industry use cases. 

\section{Related Work}

The Dialogue Breakdown Detection Challenge (DBDC) has been a pivotal platform for advancing research in this area~\cite{higashinaka-etal-2016-dialogue,hori2019overview}. \citet{higashinaka-etal-2016-dialogue} defines the task description, datasets, and evaluation metrics for DBDC and provides insights into the design and methods used in these challenges. These challenges involve detecting inappropriate system utterances that lead to dialogue breakdowns in chat, utilizing datasets composed of chat dialogues with annotated breakdown instances. The methodologies employed range from traditional machine learning techniques to advanced neural network models. \citet{Hendriksen2021} explore different variants of LSTM for dialogue breakdown detection. This work highlights the exploration of different model types and word embeddings, adding depth to the understanding of how various machine learning models and linguistic features can be utilized for breakdown detection. \citet{sugiyama2021dialogue} demonstrate a novel approach on dialogue breakdown detection by integrating BERT's powerful language understanding capabilities with traditional dialogue features like dialogue acts. This hybrid approach aims to capture the nuances of conversational flow and detect potential breakdowns more effectively.

Another significant contribution in this field is the exploration of semi-supervised learning methods to improve dialogue breakdown detection, as discussed in \citet{Ng2020ImprovingDB}. Their research demonstrates the use of continued pre-training on the Reddit dataset and a manifold-based data augmentation method, showing a substantial improvement in detecting dialogue breakdowns.  The findings across these papers consistently indicate that the integration of advanced language models with contextual and conversational features significantly enhances the detection of dialogue breakdowns.

There were a few approaches using acoustic signals or multimodality in previous related challenges~\cite{min2019predicting,Li2020MultiModalRO,kazuya2022dialog}. For more related work to our approach, \citet{meena-etal-2015-automatic} used the output from automatic speech recognition system (ASR) systems as features to detect dialogue breakdowns from spoken dialogues but they used only surface forms of STT texts or extracted text features from them rather than latent vectors of acoustic signals directly. Also, \citet{abe_2018} utilized acoustic features and found that they can classify non-breakdown dialogues and awkward conversation flows better than traditional text features but they used manually designed feature vectors extracted from emotion challenge dataset~\cite{schuller09_interspeech}.
In this paper, we explore novel multimodal architectures and the most recent state-of-the-art approaches for dialogue breakdown detection using both text and audio signals of real-time conversations in industry settings. We propose a model which uses deeper contextual signals across both audio and text inputs than prior works. This system is able to capture dialogue breakdowns in phone conversations in industry settings.

\section{Method}
\subsection{Data}
We collected our data from calls driven by our conversational AI agents to verify insurance benefits of patients for covering target medications. For our dialogue breakdown detection model training and testing, we used 1,689 phone call conversations between our AI agent\footnote{The AI agent is an independent model architecture separate from our dialogue breakdown system; we used the output of  intermediate components of this architecture as input to our dialogue breakdown detection models.} and users (e.g., insurance company employees) in which human intervention was required due to dialogue breakdowns (e.g., AI agent misclassified intent due to mistranscribed STT caused by the high level of noise during the phone calls) from August 2023. More specifically for our objective of benefit verification calls, human intervention is required when 1) AI agents do not follow standard operating procedures as in the task definition and diverge from correct paths of conversations, 2) users get frustrated during the interaction\footnote{Reasons can include repeated noncritical mistakes or slow responses of AI agents.} or 3) AI agents make critical mistakes which may cause call failures immediately or in the later phase of the calls.  We used 70\% for training, 20\% for validation and 10\% for testing for our model experiments. Then, we additionally collected 94 calls from September 2023 to test the generalizability of our best model (Table~\ref{tab:dataset}). Each phone call contains 104 to 112 turns on average between AI agents and users. The calls are randomly sampled to minimize any potential sampling bias towards specific gender, age or ethnicity of users. Binary labels of `breakdown' (turns for which human intervention was required) and `no breakdown' (turns with coherent conversation flows) are used for our dialogue breakdown detection tasks.

\begin{table}
\small
\centering
\begin{tabular}{l|l|lll}
\hline
\textbf{Dataset Type} & \textbf{Calls} & \textbf{Turns} & \textbf{AVG} & \textbf{STDV}\\
\hline
Train & 1,181 & 124,384  & 105.32  & 28.34 \\ 
Validation & 338 & 35,985  & 106.46 & 28.33 \\ 
Test & 170 & 17,690 & 104.06 & 27.63 \\
Generalizability & 94 & 10,505 & 111.75 & 34.45 \\\hline 
\end{tabular}
\caption{\label{tab:dataset} Phone call dialogue breakdown dataset. `AVG' column is the average number of turns for each call in the corresponding dataset and `STDV' column is standard deviation of turns for each call.}
\end{table}

\subsection{Models}

\begin{figure*}
    \centering
    \includegraphics[width=6in]{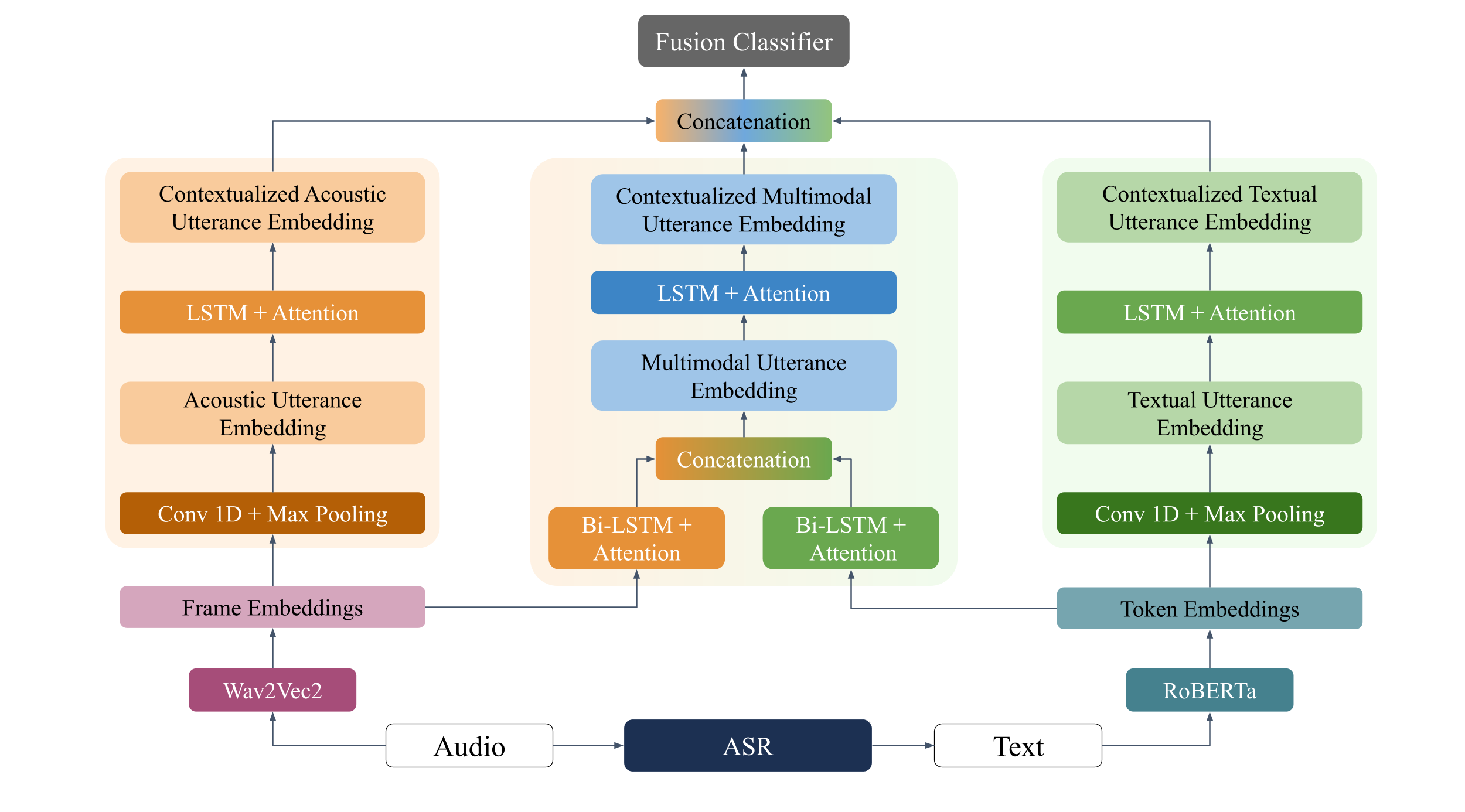}
    \caption{MultConDB model architecture. 
    }\label{fig:multimodal_lstm}
\end{figure*}
We explored potential methods including the state-of-the-art models for text only dialogue breakdown detection. For baseline, we replicated the approaches which obtained the state-of-the-art performances from the previous work: LSTM and BERT \cite{sugiyama2021dialogue}. We implemented 4 dialogue breakdown detection models that can leverage transcribed texts, and several available signals such as speaker information, intent classification of our AI model agent and raw audio signal. 


\paragraph{Text LSTM.} \label{lstm}
In this model we have extended the work of \citet{Hendriksen2021} which utilizes pre-trained GloVe embeddings to model utterance representations and use different variants of LSTM to detect dialogue breakdown~\cite{pennington2014glove}. 

In our implementation, we have extracted contextualized token embeddings using pre-trained RoBERTa \citep{liu2019roberta} model instead of non-contextualized GloVe embeddings\footnote{See more details in Section~\ref{sec:model_config:text_lstm}}. The choice of embedding is inspired by the recent surge of Transformer \citep{NIPS2017_3f5ee243} based embeddings in the literature where they are proven to yield better performance than non-contextualized GloVe embeddings. While \citet{Hendriksen2021} generates utterance embeddings by averaging all the word embeddings in an utterance, we employ a Bi-LSTM layer and attention to further process and combine the RoBERTa based token embeddings into the utterance embedding. We have employed another layer of Bi-LSTM and attention to accumulate contexts from the current and all previous utterance embeddings. The contextualized utterance embedding is then passed through a linear classifier layer, that classifies each utterance into either breakdown or non-breakdown class. 

\paragraph{End-to-End LLM Classifier.}
In the previous Text LSTM model, we used an LLM, RoBERTa as a feature extractor for extracting the token embeddings in the input utterances, but we did not use the RoBERTa model in end-to-end settings. To leverage the full capabilities of an LLM, in this LLM model we finetune the RoBERTa-base model with a classification head on top to classify the input utterance into breakdown or non-breakdown classes. In this implementation, we get rid of additional Bidirectional LSTM (Bi-LSTM) and attention networks for contextualization as we incorporate contextual information as the input to the model. Similar to the previous model, each utterance is represented as a concatenation of the speaker tag, utterance text and intent\footnote{See more details in Section~\ref{sec:model_config:e2e_llm}}. The linear layer and the layers of RoBERTa-base model are fine-tuned end-to-end. We have experimented with different configurations of the linear layers and based on the empirical results, we use 2 linear layers with 784 and 2 neurons each and the latter works as a classification layer.  

\paragraph{Multimodal Transformer (MulT A+T).}
\citet{tsai-etal-2019-multimodal} introduced the Multimodal Transformer model for emotion recognition, which leverages the transformer architecture and cross-modal attention mechanisms. This model serves as a popular baseline for emotion recognition tasks. In our research, we have adapted this model to suit the specific task of dialogue breakdown detection. The original implementation of the Multimodal Transformer incorporated three modalities (audio, video, and text) but we focus on using audio and ASR-generated text\footnote{See more details in Section~\ref{sec:model_config:mult_at}} and used positional encodings to enhance the model's understanding of positional information. 

The core component of our model is the crossmodal transformer which facilitates the integration of information from both audio and text. In the first transformer block, acoustic inputs are used as queries, while textual inputs serve as both keys and values. In the second block, these roles are reversed, with textual inputs as queries and acoustic inputs as keys and values. This approach enables effective crossmodal information exchange through attention mechanisms. Within the crossmodal transformer blocks, we employ a stack of 12 crossmodal attention layers (4 attention heads per layer). Then, the outputs are passed through traditional self-attention-based transformers (6 attention layers and 4 attention heads per layer). Following this, we apply pooling operations to the outputs from both transformer blocks and concatenate them. This concatenated representation is then passed through a series of linear layers for classification. Our architecture includes two projection layers and one classification layer. 

\paragraph{MultConDB.}
We introduce a model named \textbf{Mult}imodal \textbf{Con}textual \textbf{D}ialogue \textbf{B}reakdown detection, or MultConDB, which is inspired by and built upon the model proposed by \citet{miah-etal-2023-hierarchical} in their work on hierarchical online dialogue act classification. Our model consists of two unimodal encoder branches and one multimodal encoder branch. The unimodal encoder branches individually handle textual and acoustic features, producing two distinct unimodal encodings: one for acoustic data and one for textual data. We used Wav2Vec2~\citep{10.5555/3495724.3496768} as our acoustic feature extractor. We standardize every user or AI agent utterance by converting it into 15.0 second chunks through padding or trimming and subsequently extract frame-level features using Wav2Vec2, with each frame having a duration of 25 ms and a stride of 20 ms. For textual data, Token-level features are extracted in a similar manner, utilizing RoBERTa as described in \textbf{`Text LSTM'}.

These unimodal encoder branches share an identical architecture. Initially, frame-level or token-level features undergo processing via temporal convolutional layers, each equipped with 256 kernels (size = 5). This temporal convolution operation contextualizes the frames. Following temporal convolution, we apply a max-pooling operation to generate a single embedding vector for each utterance. We pass these utterance embeddings through an LSTM and an attention network to incorporate contextual information from a set of previous utterances. This process yields both acoustic and textual utterance embeddings.

In the multimodal branch, we employ two Bi-LSTMs along with an attention network to separately process token embeddings and frame embeddings. This approach results in a pair of utterance embeddings derived from textual and acoustic features. These embeddings are concatenated to create a multimodal embedding at the utterance level. To further enrich contextual understanding, we introduce another layer of LSTM and an attention network, taking into account context from both the current and past utterances. We empirically determine the optimal attention window size to be 5. Subsequently, we concatenate the two unimodal and one multimodal contextualized utterance embeddings to form a fusion embedding. This fusion embedding undergoes further processing through linear layers, consisting of 256 neurons and 2 neurons. Finally, the last linear layer classifies whether each input utterance is a dialogue breakdown.

\section{Results and Analysis}
We conducted dialogue breakdown classification tasks using the previous state-of-the-art models and MultConDB and analyzed MultConDB to investigate its inference process and capability with qualitative visualization analysis.  

\subsection{Task Evaluation \label{sec:task_eval}}
\begin{table}
\small
\centering
\begin{tabular}{l|l|lll}
\hline
\textbf{Model} & \textbf{Inputs} & \textbf{Prec} & \textbf{Rec} & \textbf{F1}\\
\hline
Text LSTM & S+U+I & 39.78 & 65.30 & 49.44 \\ 
End-to-End LLM & S+U+I & 64.03 & 52.35 & 57.61 \\ 
MulT A+T & S+U+I+A & 63.51 & 55.29 & 59.12 \\\hline
MultConDB & S+U+I+A  & \textbf{65.96} & \textbf{72.94} & \textbf{69.27} \\\hline
\end{tabular}
\caption{\label{tab:model_performance} Model performance for dialogue breakdown detection. Columns are defined as `Inputs': types of inputs, `Prec': precision, `Rec': recall and `F1':dialogue breakdown prediction F1 score. Each row of `Input' column values are defined as following: `S': speaker tag (\textit{AI agent} or \textit{User}), `U': utterance, `I': intent prediction of \textit{AI agent model}, `A': audio recording of utterances.} 
\end{table}
\textbf{Dialogue Breakdown Performance Analysis:} We evaluated the models in our dialogue breakdown dataset collected from August 2023 (Table~\ref{tab:dataset}). We conducted hyperparameter tuning of each model architecture using the validation set (random 20\% of the August calls) and trained with our training set (random 70\% of August calls). Each model took speaker tags, utterances, AI agent model intents of the current turn and historical turns within its context window as input and predicted whether the current turn is dialogue breakdown. We reported their performances on test set (random 10\% of August calls). We conducted fine-tuning and hyperparameter tuning of each model on August validation set.

For preliminary analysis, we used plain texts without intents predicted by our conversational AI model agent for each model architecture and in-context learning (ICL) approaches~\cite{brown2020language} with Gemini Pro and the best F1 was 40.31 (see more details in Section~\ref{sec:prelim}). This result suggested that it might be difficult to capture phone call dialogue breakdowns within a few shots without audio context such as noise or intonation or voice tones of users. Thus, we first explored text based models by fine-tuning and training with dialogue breakdown turns as labels so they can leverage the full training dataset (`Text LSTM' and `End-to-End LLM'). Then, we trained multimodal models with both acoustic and text signals (MulT A+T and MultConDB). 

In general, multimodal models obtained higher F1 scores than text only models. This trend suggests that multimodal settings leveraging both text and audio signals are more effective for capturing phone call dialogue breakdowns. Among Multimodal models, MultConDB obtained the best F1 score for classifying dialogue breakdowns 15\% F1 score improvement over Multimodal Transformer (p < 0.001). This may indicate that multimodal contextual model architecture and training process designed specifically for detecting dialogue breakdown are critical to obtain the performance level of practical use.

\textbf{False Positive Analysis:} In addition to detecting exact dialogue breakdown turns, it is also important for models to make false positive dialogue breakdown predictions as near as possible to the actual breakdown down point if any. In industry settings in which a large scale automated phone calls concurrently happen, the dialogue breakdown detection model should bring human in the loop near the breakdown points otherwise human feedback at the false positive turns are not useful as well as overall call failure rates may increase because other phone calls with actual dialogue breakdown may not get support from human intervention in time. 

\begin{figure}
    \centering
    \includegraphics[width=3in]{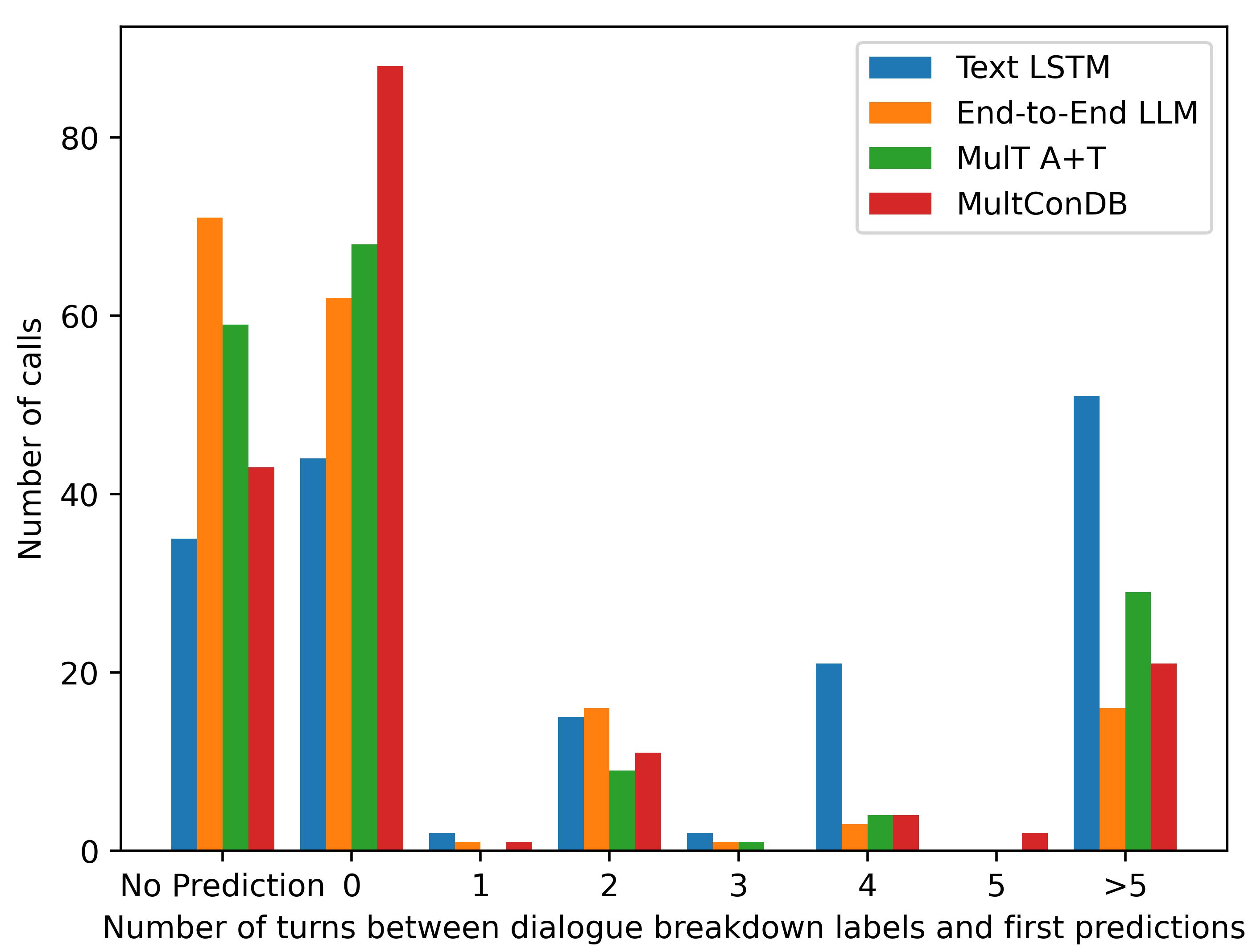}
    \caption{Number of turns between dialogue breakdown ground truth and first model predictions.}
    \label{fig:false_positive_analysis}
\end{figure}

In Figure~\ref{fig:false_positive_analysis}, we measured the number of turns between the dialogue breakdown turns and the first dialogue breakdown predictions of the models. Multimodal models tend to make dialogue breakdown predictions in nearer turns to breakdowns than text only models do. This difference might have been caused by audio context which can provide noise or speech timing of user and AI agent which might cause dialogue breakdown in the near future. Although End-to-End LLM had the smallest number of false positives more than 5 turns away from the breakdown points, it has the largest number of false negatives (`No prediction' in Figure~\ref{fig:false_positive_analysis}) and this trend can increase overall dialogue breakdown detection failure rates. MultConDB obtained the highest of true positive predictions and the second lowest number of false positives more than 5 turns away and false negatives following the extreme high precision model (End-to-End LLM) and high recall model (Text LSTM) respectively.

\subsection{MultConDB Qualitative Analysis}

\begin{figure}
    \centering
    \includegraphics[width=3in]{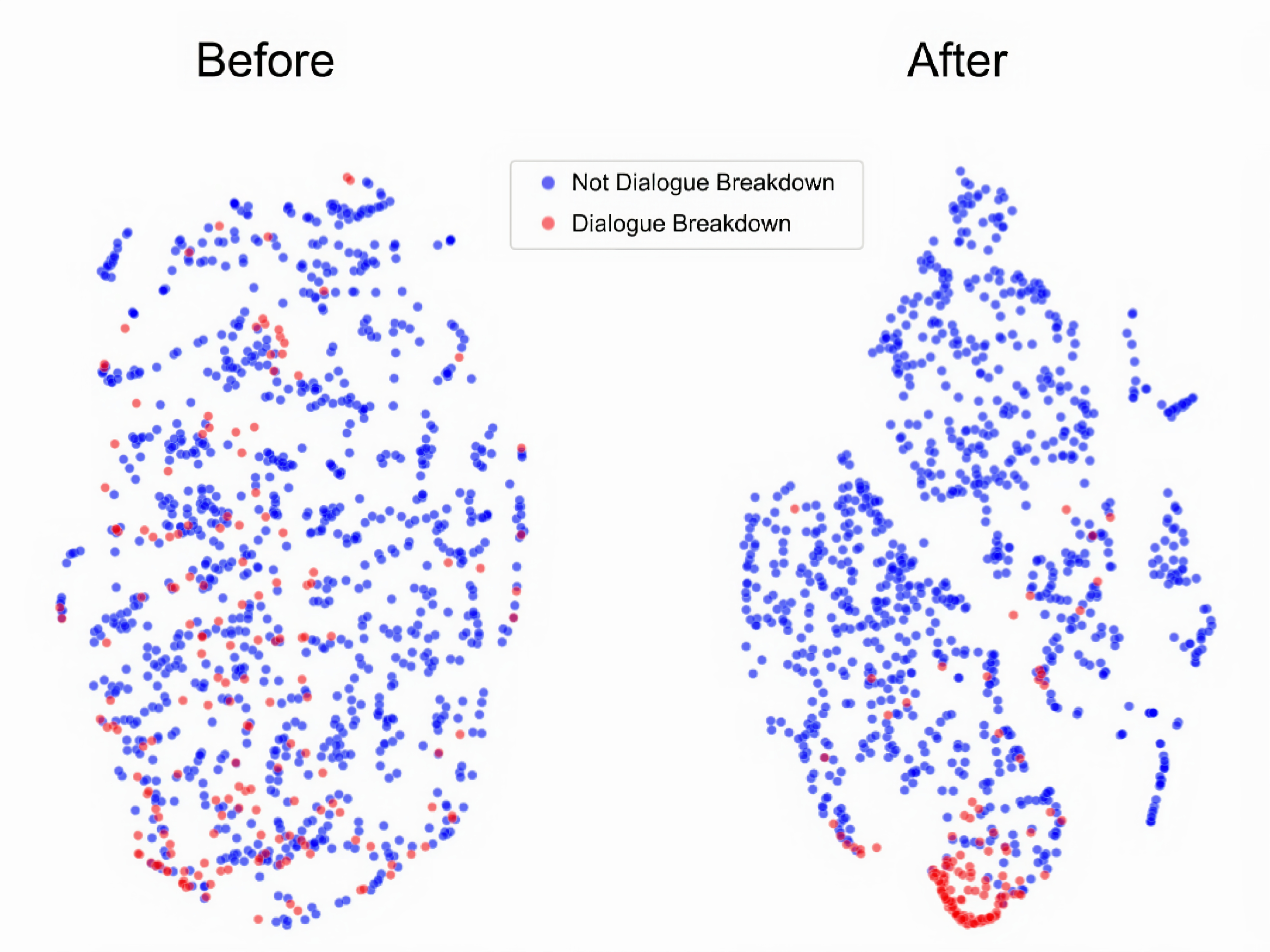}
    \caption{Breakdown and non-breakdown turns of users and our conversational AI model captured by our model output layer. \textit{Before} figure shows 2D t-SNE of our model input embedding (concatenation of speaker tag, utterance, AI agent intent and audio) and \textit{After} figure shows the last output layer of our model right before prediction head softmax layer. 
    }
    \label{fig:model_output_layer_analysis}
\end{figure}

\textbf{Model Output Analysis:} We analyzed the efficacy of our dialogue breakdown detection model in terms of whether our multimodal contextual model architecture is effective for capturing phone call dialogue breakdowns.
In Figure~\ref{fig:model_output_layer_analysis}, model inputs in \textit{Before} figure show that breakdown and non-breakdown turns are not linearly separable and difficult to be classified without contextualization; they are spread around without any specific patterns. In contrast, our model outputs in \textit{After} suggest that our model architecture was quite effective for all types of dialogue breakdown turns; breakdown turns are clustered in the most right side of the figure. Dialogue breakdown turns are difficult to be captured because the same utterance text can be a dialogue breakdown turn or natural conversation flow turn based on the context and acoustic signals. Also, the same sentence can be a question, statement, or continuing speech with a short pause before the following statement based on its intonation and context so the response of our AI agent is likely to cause dialogue breakdown and provide negative experience to users if it interrupts users' speech or it goes silent when it misunderstood a question as a statement. This analysis suggests that  MultConDB can classify dialogue breakdown utterances leveraging these types of subtle nuances and contexts. 

\begin{figure}
    \centering \includegraphics[width=3in]{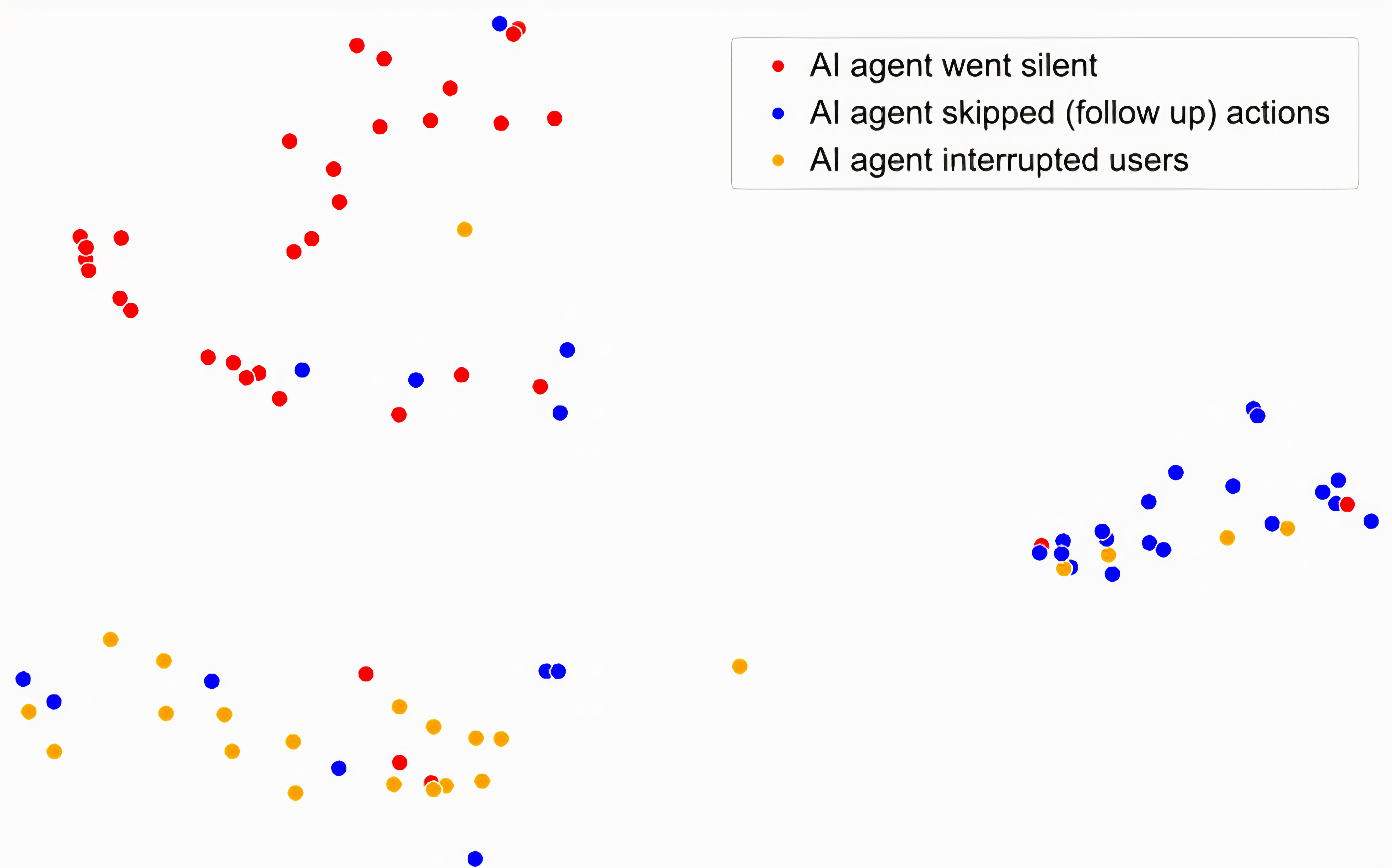}
    \caption{2D t-SNE of MultConDB output layers colored by types of dialogue breakdown.}
    \label{fig:error_categorization}
\end{figure}

\textbf{Underlying Causes of Breakdowns:} For additional analysis to validate whether MultConDB is effective for inherently categorizing types of dialogue breakdowns further, we conducted a visualization analysis for MultConDB output layers for its capability of categorizing the causes of dialogue breakdown. Among dialogue breakdown utterances in our testset phone calls, we identified the most distinguishable and clear causes of dialogue breakdowns as following: AI agent went silent (34 turns)\footnote{For example, AI agent may wait for responses when it misclassifies the end of speech from users as continuing speech or upstream STT finalization was delayed (Figure~\ref{fig:dialogue_breakdown_example1_ai_agent_goes_silent}).}, AI agent interrupted users (23 turns)\footnote{For example, AI agent may misclassify continuing speech of users as the end of speech and ask a next question while the users are providing their answers (Figure~\ref{fig:dialogue_breakdown_example1_ai_agent_interruption}).}, and AI agent skipped required actions or follow up actions (31 turns)\footnote{For example, loud noises can cause a high rate of STT mistranscriptions which cause intent misclassfication of AI agent (Figure~\ref{fig:dialogue_breakdown_example}).}. 
Although we have not trained MultConDB with explicit labels of types of dialogue breakdown, it inherently captured which type of underlying causes led to dialogue breakdown. In Figure~\ref{fig:error_categorization}, the turns after which AI agent went silent were clustered on the top left and the turns in which AI agent interrupted the speech of users were clustered on the bottom left. Finally, the turns where AI agent skipped required actions or follow up actions were clustered on the right side. This suggests that MultConDB can capture abnormal conversation flows based on acoustic and text contexts such as the voices of AI agents and users are combined in one turn\footnote{Potential acoustic signals of AI agents interrupting users.} or AI agent not following up after the question intonation of users' speech breaking the alternating turns of each voice.

\subsection{Dialogue Breakdown Detection Model Generalizability Testing} \label{sec:generalizability_main}
\begin{figure}
    \centering \includegraphics[width=3in]{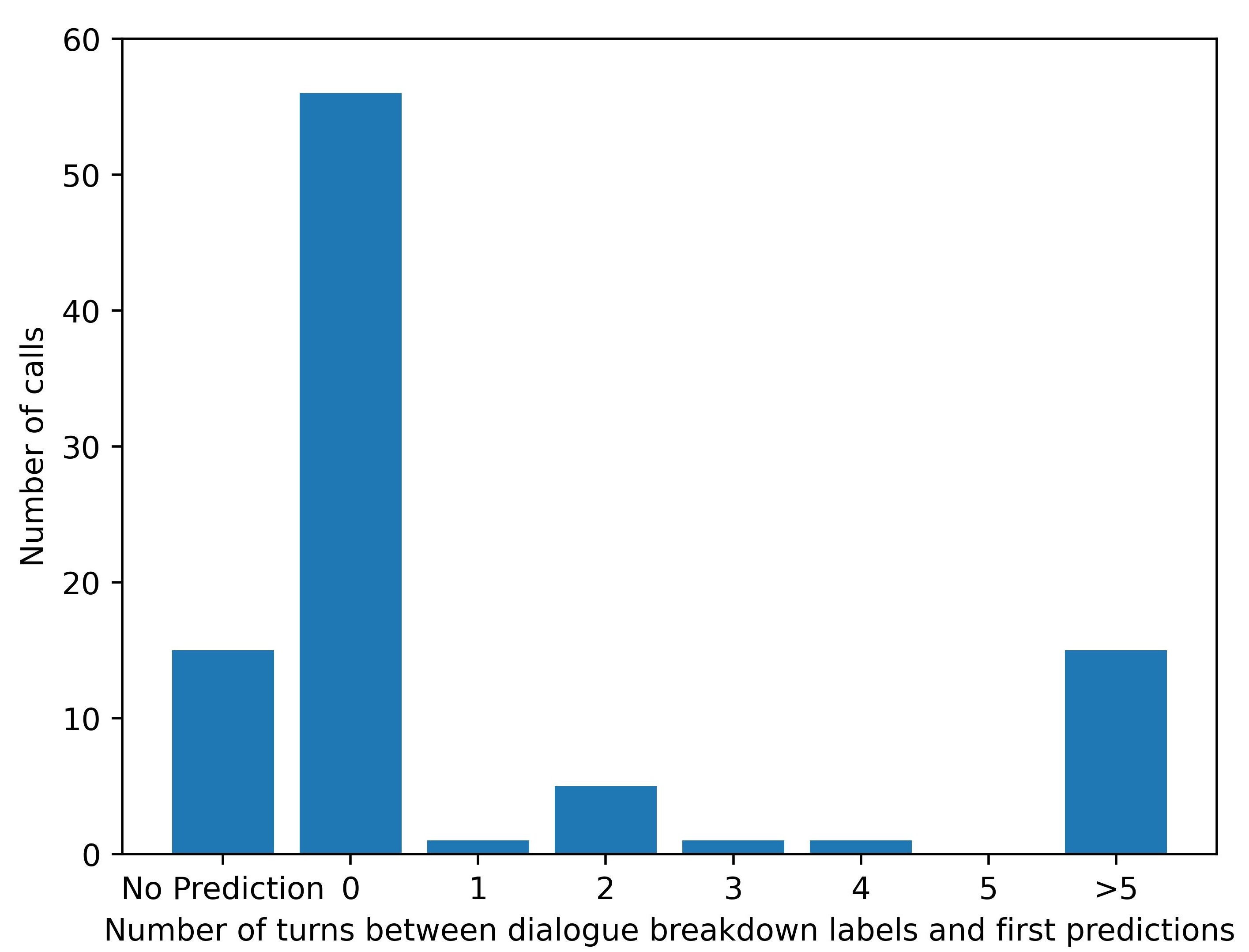}
    \caption{Number of turns between dialogue breakdown
ground truth and first model predictions in September 2023 calls.}
    \label{fig:generalizability}
\end{figure}

In real-world industry settings, dialogue flow and utterance distributions in phone calls keep changing everyday because outbound call target users have different call volumes everyday and they may ask different types of questions based on their policy changes. Also, the conversational AI agent models are maintained and updated with new releases. Thus, it is critical for the dialogue breakdown models to generalize effectively to various types of dialogue flows and interactions which have not been observed during its training process. 
To that end, we tested our dialogue detection model on unseen data. We used 94 calls from September 2023 (Table~\ref{tab:dataset}) which include calls driven by updated AI agent models with new releases to validate how well MultConDB can generalize to unseen calls. MultConDB obtained F1 score 71.22 with precision 65.77 and recall 77.66 maintaining a high performance. In Figure~\ref{fig:generalizability}, it had an almost similar pattern to its prediction pattern from its August 2023 call data although the model is used for classifying dialogue breakdown calls in a different month. Even though it had a relatively high number of false positive predictions more than 5 turns away from the actual dialogue breakdown differently from the call data August, it still had the largest number of true positive predictions (0 in the figure) followed by no prediction as the second highest number of prediction category. These results suggest that MultConDB was not biased towards the types of dialogue breakdowns caused by the previous release version of AI agent model and it can generalize well to other types of dialogue breakdowns including the in correct reponses or abnormal conversation flows caused by the updated version of AI agent model or potentially other types of context introduced in a different month.

\section{Conclusion}
As the first work of multimodal dialogue breakdown detection in healthcare industry settings, we explored various approaches leveraging both audio and text and developed a high-performance multimodal model (F1 = 69.27) which generalized well to various types of context of conversations driven by conversational AI models across multiple version releases in industry settings. Additionally, we conducted thorough qualitative model analysis which provided insights with its output patterns which can cluster dialogue breakdown samples and their categories in separate groups. We hope the strong results of our dialogue detection approach here leads to more reliable conversational AI model development in the future research. 

\section*{Limitations}
Due to PHI concern, we cannot make our dataset publicly available and we explored model architectures which can be locally hosted instead of API calls.

\bibliography{custom}

\appendix
\section{Phone Call Dialogue Breakdown Examples}
\label{sec:db_example}
\begin{figure}
    \centering
    \small
    \includegraphics[width=3in]{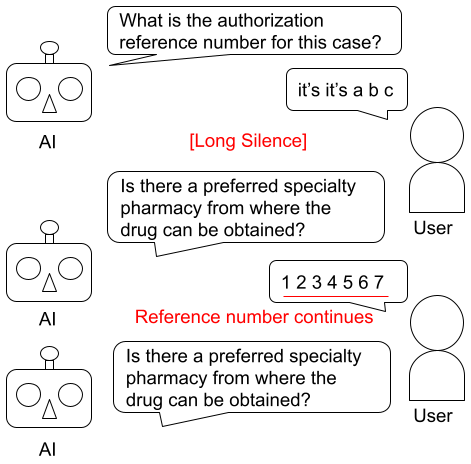}
    \caption{AI Agent misunderstood a partial sequence `abc' followed by a relatively long pause from the user as a full reference number and interrupted the user by asking its next question in the middle of the user speech.
    }
    \label{fig:dialogue_breakdown_example1_ai_agent_interruption}
\end{figure}

\begin{figure}
    \centering
    \small
    \includegraphics[width=3in]{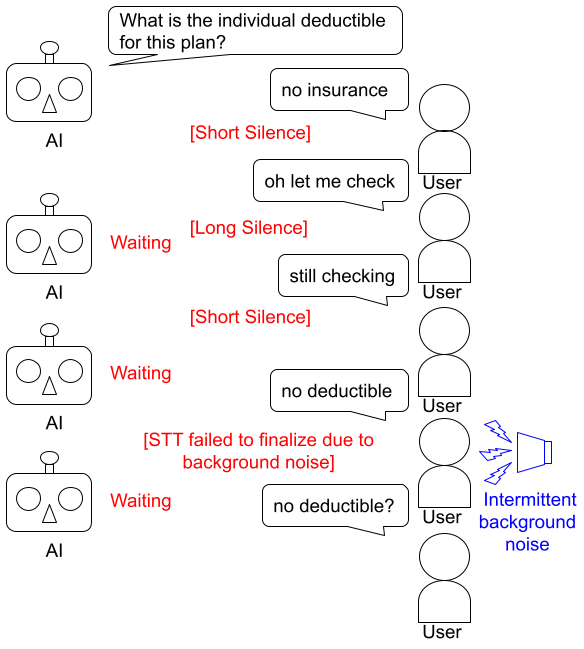}
    \caption{AI Agent went silent even when the user provided an answer due to the combined factors of ASR STT finalization, downstream intent misclassification.}
    \label{fig:dialogue_breakdown_example1_ai_agent_goes_silent}
\end{figure}

\begin{figure}
    \centering
    \small
    \includegraphics[width=3in]{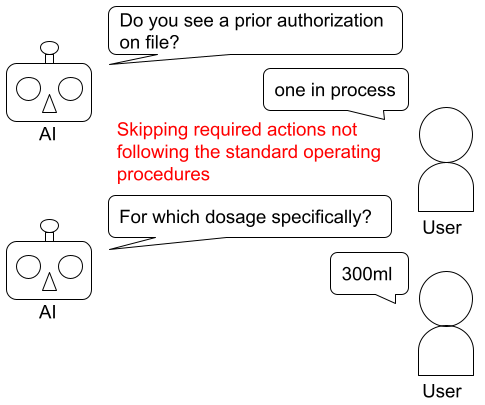}
    \caption{AI Agent should have asked if there is an active prior authorization which can cover target medication when the user said `one in process' instead of proceeding with the pending prior authorization asking its next question.}
    \label{fig:dialogue_breakdown_example1_ai_agent_skipping_actions}
\end{figure}

Differently from dialogue breakdown examples from interactions between users and text only chatbot AI agents, conversation flows of dialogue breakdown from our phone call data contain various types of examples with a higher complexity caused by additional factors such as audio-related issues, real-time conversation latency expectations and health insurance benefit verification standard operation procedures. For example, users can take various lengths of pause between their utterances while they provide long sequences of information (e.g., phone numbers, patient ID, processor control number or bank identification number) so AI agents may confuse short pauses of users in the middle of the full sequence with an end of speech and start asking next questions (Figure~\ref{fig:dialogue_breakdown_example1_ai_agent_interruption}). The variance of speech pace and lengths of pauses from users can confuse AI agents and aggravate the situation even further when the user speech contains multiple confusing utterances in a row due to potentially various underlying factors such as incorrect STT finalization which may cause intent misclassification in the downstream pipeline of AI agents (Figure~\ref{fig:dialogue_breakdown_example1_ai_agent_goes_silent}). More complex examples include more subtle nuanced situations in which AI agents apparently drove conversations correctly but missed required actions which may lead to call failure in the later phase of the call (Figure~\ref{fig:dialogue_breakdown_example1_ai_agent_skipping_actions}).

\section{Preliminary Analysis}
\label{sec:prelim}
We conducted preliminary analysis using the latest state-of-the-art conversational AI models to validate the task, dataset and level of difficulty for dialogue breakdown model development.

\subsection{Gemini Pro In-Context Learning}
We used in-context learning (ICL) approaches with Gemini Pro as this model did not support fine-tuning and selected Gemini over other LLM vendors due to the status of security review at the time to remain HIPAA compliant. We provided randomly selected N calls from training set with the prompt in the following format\footnote{We designed this prompt and experiment settings based on the characteristics of our phone call conversation dataset and the previous ICL exploration work~\cite{brown2020language,min2022noisy,logan2022cutting}}: 
\\ \\
\noindent\fbox{%
    \parbox{2.9in}{%
        Given a conversation between an AI agent and a user, find a dialogue breakdown turn which may need a human intervention.

The conversation is in the following format "(turn number)[AI Agent]: what the AI Agent said|(turn number)[User]: what the user said".
Return which turn needs human intervention due to dialogue breakdown. 
For example, from the conversation input "(1)[User]: How are you today|(2)[AI Agent]:I want to eat ice cream", you can return 2 because the second turn was off-the-topic.

Which turn do you think may cause a dialogue breakdown so it may need some human intervention from the following conversation?

Here are some real phone call conversation examples:

\{ Example N calls randomly selected from training set in the given conversation format along with their ground truth breakdown turns \}

Now, provide your answer from this conversation:

\{Testset call in the given conversation format\}

    }
}

\qquad \\
The largest number of sample calls were used as examples in context as long as Gemini Pro context limit allows (32,000 tokens); prompts with 33 call or more examples caused `invalid argument' errors based on the lengths of sampled calls. We used temprature = 0, top\_p=1, top\_k=40, candidate\_count=1 and max\_output\_tokens=8000\footnote{We used a large number of output tokens because Gemini may provide long answers such as `There is no dialogue breakdown in this conversation.' or `The conversation does not contain the answer to this question.'} so the results can be reproducible. We conducted random sampling for ICL call examples with random seed of default value (None) and from 0 to 9 and reported mean, 25\% and 74\% percentile F1 performances of 11 iterations of each number of example calls. The highest mean F1 score of Gemini Pro PCL was 40.31 when it had 31 random example calls and this number was reported in our main paper (Section~\ref{sec:task_eval}). 

The general trend of F1 score was increasing up to this point with a few decreasing patterns in the middle and this general trend aligns well with the findings from related prior work~\cite{Zhao2021CalibrateBU,liu-etal-2022-makes,min2022rethinking}. However, Gemini does not have explicitly encoded information for the healthcare industry phone calls from our datasets so its performance tend to fluctuate based on which calls are sampled as context examples for detecting dialogue breakdowns from the given test calls.

\begin{figure*}
    \centering \includegraphics[width=6in]{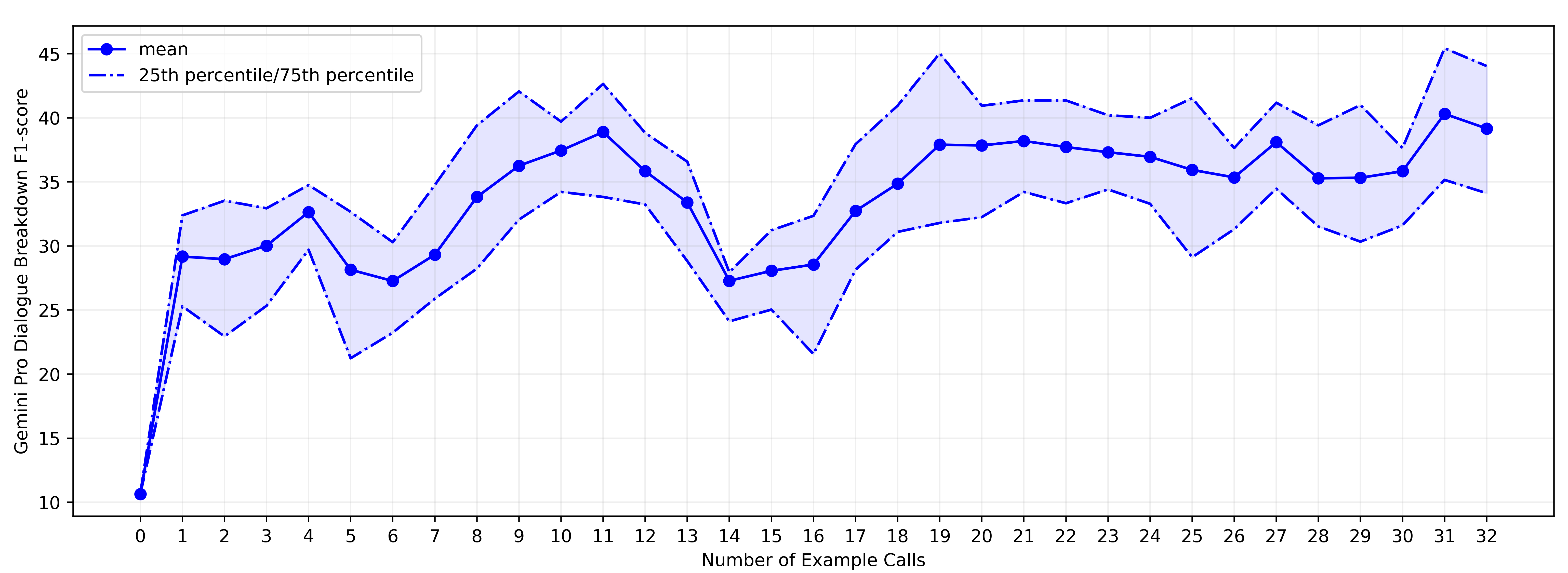}
    \caption{Gemini Pro in-context learning dialogue breakdown prediction performance changes among the number of example calls provided in context.}
    \label{fig:gemini_pro_ICL}
\end{figure*}

\section{Model and System configurations}
\label{sec:model_config}
\subsection{Text LSTM}
\label{sec:model_config:text_lstm}
\paragraph{Input Processing.} Each utterance is represented in the format -  `speaker: \textit{speaker tag (AI agent or user)} | utterance: \textit{utterance text generated by ASR} | intent: \textit{intent of the utterance from AI Agent input classification model}' and passed as the input to the feature extractor model.
\subsection{End-to-End LLM}
\label{sec:model_config:e2e_llm}
\paragraph{Input Processing.} Each input to the model is represented in the format - `<s> \textit{current utterance representation} </s> \textit{four previous utterance representations} </s>.' Then we join the current utterance representation and the four previous context utterance representations with separator token. The input is passed to the RoBERTa-base model and we extract the embedding of the sentence start token, <s> as the utterance embedding and pass it to a set of linear layers for classification.
\subsection{MulT A+T}
\label{sec:model_config:mult_at}
\paragraph{Input Processing.} Specifically, we employ RoBERTa as the textual feature extractor, following a similar process as described in \ref{lstm} to extract token embeddings. For acoustic features, we utilize Wav2Vec2 \citep{10.5555/3495724.3496768} to extract information from the raw waveform data. To ensure consistency, we pad or trim every utterance signal to a fixed duration of 15.0 seconds. After extracting the modality-specific features, we apply two 1D convolutional layers, each consisting of 256 kernels with a size of 5, to make the input features aware of their temporal neighborhood.
\subsection{MultConDB} \label{sec:model_config:multcondb}

\paragraph{Hyperparameters.} To ensure the reliability of the results, each experiment is carried out using three different seeds. The primary metric for evaluating the results is the F1 score. The training process continue for 40 epochs, with an early-stopping mechanism implemented to stop training if there is no improvement in the F1 score for five consecutive epochs. Table \ref{tab:hyperparameters} displays a detailed list of hyperparameters along with the values selected based on experimentation.

\begin{table}
\small
\centering
\begin{tabular}{|l|l|}
\hline
\textbf{Name} & \textbf{Best Value} \\
\hline
Batch size & 32\\
Hidden dimensions of encoders & 128\\ 
Kernel size of conv layers & 5\\
No of channels of conv layers & 256\\
Window size for context modeling & 5\\
\hline
\end{tabular}
\caption{Choice of hyperparameters}
\label{tab:hyperparameters}
\end{table}

\paragraph{Design Choices.} In our implementation, we use Wav2vec2 as the acoustic feature extractor instead of Whisper which is used as a feature extractor in \citet{miah-etal-2023-hierarchical}. However, Whisper \citep{radford2023robust} requires audio chunks to be adjusted to a fixed length of 30 seconds, which significantly exceeds the typical duration of utterances in our dataset. Wav2Vec2 offers flexibility in terms of audio chunk length, allowing us to use 15 second chunks. This not only makes our model more memory-efficient but also speeds up the inference process to meet the demands of our online task. To improve the model's online performance, we opt to restrict the context size to 5. Through experimentation with various context lengths during inference, we observe that while a larger context typically yields better performance, it also increases the inference time. In our design, we use a context length of 5, which strikes a balance between achieving satisfactory performance and maintaining fast inference speed. The inference time is 0.06s for each utterance.

\end{document}